\documentclass[conference]{IEEEtran}
\IEEEoverridecommandlockouts
\usepackage{cite}
\usepackage{amsmath,amssymb,amsfonts}
\usepackage{algorithmic}
\usepackage{graphicx}
\usepackage{textcomp}
\usepackage{xcolor}
\usepackage{xurl}
\usepackage[hidelinks]{hyperref}
\usepackage{eso-pic}

\usepackage{soul}
\def\BibTeX{{\rm B\kern-.05em{\sc i\kern-.025em b}\kern-.08em
    T\kern-.1667em\lower.7ex\hbox{E}\kern-.125emX}}
\begin{document}

\AddToShipoutPictureFG*{%
  \AtPageUpperLeft{%
    \put(0,-34){\makebox[\paperwidth]{%
      \parbox{0.86\paperwidth}{\centering\scriptsize
        \copyright~2026 IEEE. Personal use of this material is permitted. Permission from IEEE must be obtained for all other uses, in any current or future media, including reprinting/republishing this material for advertising or promotional purposes, creating new collective works, for resale or redistribution to servers or lists, or reuse of any copyrighted component of this work in other works.\\
        Accepted for publication in Proc.\ 48th Annual International Conference of the IEEE Engineering in Medicine and Biology Society (EMBC 2026).
      }}}}}

\title{Cross-Linguistic Acoustic Biomarkers of Cognitive Impairment Across Interactive and Single-Speaker Speech Tasks
}

\author{
\IEEEauthorblockN{
Kenyu Ikeuchi\IEEEauthorrefmark{1}, 
Tomoaki Ohtsuki\IEEEauthorrefmark{1}, 
Michelle Cohn\IEEEauthorrefmark{2}, 
Alyssa Weakley\IEEEauthorrefmark{3}, 
Yui Ishihara\IEEEauthorrefmark{2},\\
Joseph Hoang\IEEEauthorrefmark{2},
Taishiro Kishimoto\IEEEauthorrefmark{4},
Sarah Tomaszewski Farias\IEEEauthorrefmark{3},
Jet M.J. Vonk\IEEEauthorrefmark{5},
and Chen-Nee Chuah\IEEEauthorrefmark{2}
}
\IEEEauthorblockA{\IEEEauthorrefmark{1}Keio University, Yokohama, Japan}
\IEEEauthorblockA{\IEEEauthorrefmark{2}University of California - Davis, Davis, CA, USA}
\IEEEauthorblockA{\IEEEauthorrefmark{3}University of California - Davis Health, Sacramento, CA, USA}
\IEEEauthorblockA{\IEEEauthorrefmark{4}Keio University School of Medicine, Tokyo, Japan}
\IEEEauthorblockA{\IEEEauthorrefmark{5}University of California - San Francisco, San Francisco, CA, USA}
}
\maketitle

\begin{abstract}
Multilingual dementia detection is essential for scalable and equitable screening. In recent years, automated speech analysis has emerged as a promising approach for supporting dementia detection by enabling non-invasive and scalable assessment of cognitive function. However, existing studies that use automated speech analysis often rely on one type of task, such as picture description, and use black-box models. These studies often neglect the role of task and language. To address these gaps, we performed a cross-linguistic and cross-task acoustic analysis that integrates a conversational dataset with structured speech tasks across four languages (Spanish, Mandarin, English, Japanese). We applied statistical, linguist-/clinician-guided, and machine learning model-based feature selection to derive compact acoustic feature sets while maintaining classification performance. This enabled systematic comparison of cognitive impairment-relevant features across datasets and tasks. Our results revealed that acoustic features did not generalize uniformly across tasks and languages. Temporal features remained robust across all datasets, consistently capturing timing disruption regardless of language or task. In contrast, speech rate showed strong task dependence—emerging as informative mainly in single-speaker tasks with externally anchored content and pacing, but becoming less reliable when speech content was highly individualized. Pitch-related features further exhibited language-specific patterns, favoring chroma-related features in tonal languages (e.g., Mandarin) and F0-related features in non-tonal languages. These findings caution against indiscriminate pooling of heterogeneous speech data and underscore the need for interpretable, context-aware approaches to multilingual dementia detection using automated speech analysis.

\end{abstract}


\section{INTRODUCTION}
Dementia and Mild Cognitive Impairment (MCI) constitute major global health challenges, currently affecting more than 55 million individuals worldwide, with prevalence expected to rise substantially due to population aging \cite{who_dementia}. Clinical diagnosis typically relies on neuropsychological assessments, which require trained specialists and are influenced by linguistic, cultural, and educational factors. These constraints limit scalability, particularly in underserved regions where access to specialized clinical resources remains scarce.

Speech-based analysis offers a promising complementary approach, as language can be collected non-invasively using accessible devices such as smartphones, enabling scalable screening and longitudinal monitoring~\cite{Yamada2023MobileAD}. While transcript-based methods show strong diagnostic potential~\cite{Ding2024AISurveyAD}, they omit paralinguistic cues such as vocal quality and prosody. In contrast, direct acoustic analysis preserves production-level information and avoids reliance on transcription, which is particularly beneficial for pathological or atypical speech~\cite{Niemela2025DementiaAcoustic}.

While the majority of adults around the world speak more than one language~\cite{UNESCO2024Multilingual}, global deployment of speech-based dementia screening remains challenging due to severe cross-linguistic data imbalance. This imbalance arises because validated datasets are concentrated in a small number of high-resource languages, such as English~\cite{Shakeri2025MultiConAD}. This has motivated increasing interest in multilingual dementia detection.
In particular, prior studies have increasingly relied on large pre-trained transformer models~\cite{jia25_interspeech, melistas23_interspeech}. While these studies achieve strong classification performance, important limitations remain.

Specifically, studies to date have predominantly focused on structured, single-speaker tasks such as picture description (PD) (describing everything they see in an image)~\cite{luz21_interspeech} or reading a passage aloud~\cite{ivanova21_csl}. In contrast, spontaneous conversation—despite its clinical relevance (e.g., clinician visits)—remains underexplored in dementia research. In addition, while different speech tasks impose distinct cognitive and interactional demands,  task-related variation is often overlooked. Moreover, transformer-based embedding approaches provide limited clinical interpretability and largely do not integrate expert knowledge from linguistics and clinical practice, making it difficult to relate model decisions to clinically meaningful speech behaviors.

To address these limitations, this study conducts a cross-linguistic and cross-task acoustic analysis across five heterogeneous datasets spanning four languages and four speech tasks, including a single conversational dataset. Rather than prioritizing classification accuracy, we emphasize interpretability and clinical relevance by analyzing select acoustic features, guided by linguistic and clinical dementia diagnosis experts.

The main contributions of this study are as follows:
\begin{itemize}
    \item We present a cross-linguistic and cross-task acoustic analysis 
    across five datasets spanning four languages and four speech tasks, 
    including a conversational dataset.

    \item We propose an interpretable acoustic analysis framework guided by linguistic and clinical expertise.

    \item We systematically examine how language and task modality 
    shape cognitive impairment (CI)-related acoustic markers.
\end{itemize}
\section{RELATED WORK}

Recent multilingual dementia detection research has been strongly influenced by large pre-trained transformer models, which aim to improve cross-lingual robustness via large-scale multilingual speech representations. Broadly, existing studies fall into two categories based on data usage: (i) multilingual data pooling and (ii) cross-lingual transfer learning.

\subsection{Multilingual Data Pooling Approaches}

A dominant line of work performs multilingual dementia detection by pooling speech data from multiple languages to learn language-agnostic features. For example, Jia et al.~\cite{jia25_interspeech} propose multilingual fine-tuning of Whisper to improve overall detection performance. Similarly, Favaro et al.~\cite{favaro25_embc} develop a multilingual and multimodal framework integrating language-agnostic embeddings, language-dependent components, and demographic information, demonstrating improved generalization across English, Spanish, and Mandarin Chinese. In a related setting, Agbavor et al. \cite{agbavor24_brainsci} compare language-agnostic and language-specific pipelines using Whisper embeddings on English and Mandarin Chinese speech, achieving strong multilingual performance via ensemble modeling.

\subsection{Cross-Lingual Transfer Learning Approaches}

Recent studies evaluate cross-lingual generalization under explicit train--test language mismatches. These studies assess whether dementia-related speech markers learned in one language transfer to unseen target languages. For instance, Melistas et al. \cite{melistas23_interspeech} investigate zero-shot and few-shot transfer by training on English and evaluating on Greek and Spanish. Similarly, Chen et al. \cite{chen23_icassp} examine English-to-Greek transfer by combining paralinguistic features with pre-trained acoustic embeddings.

\subsection{Positioning of the Present Study}

While prior work either emphasizes multilingual performance via data pooling or evaluates cross-lingual transfer in isolation, the role of task modality and interactional structure remains underexplored. Furthermore, relying on opaque embedding representations limits interpretability.

Motivated by these gaps, the present study adopts a complementary approach by focusing on interpretable acoustic features and explicitly analyzing how language and task modality shape cognitive impairment-related speech markers, including spontaneous and conversational speech, across five datasets.

\section{METHODS}

\subsection{Datasets}

This study examines five datasets of speech produced by older adults varying in cognitive status in English, Spanish, Mandarin Chinese, and Japanese. 
We adopt the original clinical diagnostic labels provided in each dataset, categorizing participants into healthy controls (HC) and cognitively impaired (MCI / Dementia) groups. This approach respects the comprehensive medical assessments performed by clinicians, using the Mini-Mental State Examination (MMSE)—a brief cognitive screening test~\cite{arevalo15_cochrane}—only as a supplementary screening measure rather than a definitive diagnostic standard. To improve cross-dataset comparability (as some datasets include only MCI but not Alzheimer’s disease), we excluded recordings from individuals with severe cognitive impairment (MMSE $<$ 10). The final composition of each dataset is detailed in Table~\ref{tab:dataset-composition}.

\begin{table}[t] 
\centering 
\caption{Dataset composition based on clinical diagnosis (after excluding MMSE $<$ 10). }
\label{tab:dataset-composition}
\begin{tabular}{lcc} \hline 
\textbf{Dataset (Lang, Task)} & \textbf{Normal}& \textbf{Impaired} \\ \hline 
ADReSSo (English, PD) & 114 (HC) & 116 (AD)  \\
Ivanova (Spanish, Reading) & 150 (HC) & 127 (AD, MCI) \\
Chou (Mandarin, PD) & 43 (HC)& 44 (MCI) \\
PROMPT (Japanese, Conv) & 74 (HC) & 52 (CI) \\
ADRC (English, AM) & 98 (HC) & 110 (MCI, Dementia) \\ \hline 
\end{tabular} 
\end{table}

\subsubsection{ADReSSo} The ADReSSo dataset~\cite{luz21_interspeech} is a benchmark collection of older adults’ spontaneous speech describing the Cookie Theft picture in English, sourced from individuals with Alzheimer’s disease (AD) and cognitively healthy controls (HC). 

\subsubsection{Ivanova} The Ivanova dataset~\cite{ivanova21_csl} is a Spanish corpus of reading-aloud speech in which participants recite the opening sentences of Cervantes’ \textit{Don Quixote}. This dataset includes speakers diagnosed with AD, MCI, and HC. 

\subsubsection{Chou} The Chou corpus~\cite{macwhinney2011aphasiabank} is a Mandarin Chinese speech dataset collected in Taiwan, and constitutes the Mandarin subset of the TAUKADIAL benchmark dataset~\cite{luz24_interspeech}. Participants (MCI and HC) performed three picture-description tasks (PD), and features were extracted from each recording and averaged across the three picture-description tasks. 

\subsubsection{PROMPT} The PROMPT dataset~\cite{kishimoto20_prompt} is a Japanese clinical corpus in which participants—including individuals with cognitive impairment (CI) and cognitively healthy controls (HC)—engaged in a brief free conversation (Conv) with a trained psychiatrist. Although multimodal information is available, this study uses only the audio from the free conversational segment. We retained a single recording per individual to avoid repeated measures. 

\subsubsection{Davis ADRC} The University of California Davis Alzheimer's Disease Research Center (ADRC) dataset consists of prompted spontaneous speech collected from older adults across the cognitive spectrum (Dementia, MCI, HC)~\cite{adrc}. Interviewers elicited extended narrative responses by asking participants to describe personal life events—specifically a favorite vacation, their wedding (if married, otherwise their first job), and their hometown. These prompts are designed to encourage natural speech to engage autobiographical memory (AM). Although recordings in the Spanish language are available in the dataset, their sample size is substantially smaller than that of the English recordings, which may compromise analytical stability.
Therefore, only the English recordings were used in the present analysis. 

\subsection{Overview of the analysis pipeline}

\begin{figure*}
    \centering
    \includegraphics[width=1.0\linewidth]{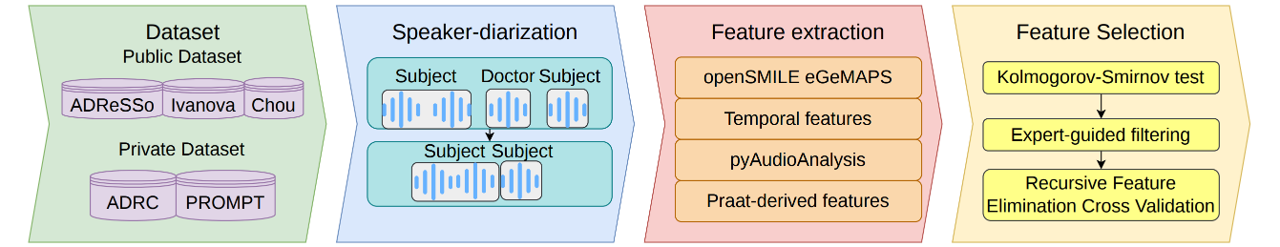}
    \caption{Overview of the proposed method}
    \label{fig:pipeline}
\end{figure*}

An overview of the proposed analysis pipeline is illustrated in Fig.~\ref{fig:pipeline}.
The workflow consists of four main stages: (i) acoustic feature extraction from speech, (ii) statistical feature filtering, (iii) expert-guided filtering, and (iv) model-based feature selection.

\subsubsection{Feature extraction}
All recordings were processed with a pyannote.audio~\cite{bredin2020pyannoteaudio} speaker-diarization pipeline to obtain frame-level speaker labels.
This step produced two outputs: (i) temporal timestamps marking the start and end of each participant's utterance, and (ii) a concatenated audio signal containing only the participant's speech (i.e., excluding an experimenter/interviewer). 
The concatenated signal was used for acoustic feature extraction using established and widely adopted toolkits (openSMILE eGeMAPS~\cite{eyben2010opensmile}, pyAudioAnalysis~\cite{giannakopoulos2015pyaudioanalysis}, and Praat Acoustic Analysis Scripts~\cite{deJong2009Praat}), which have been used in prior speech and dementia research.

Table~\ref{tab:feature_counts} summarizes the resulting feature groups and their dimensionalities for single-speaker and conversational tasks.

\subsubsection{Temporal features} Temporal features were derived from the diarized segments. For each participant utterance, we computed 7 summary statistics (sum, minimum, maximum, median, mean, standard deviation, and count) of the duration. Additionally, speech activity was quantified as the global speech ratio (participant speaking time / total recording duration). To capture interaction dynamics, we classified silent intervals based on speaker turns. We defined \textit{medial pauses} as silence between consecutive utterances within the participant's turn. For conversational tasks involving an interviewer, we additionally extracted \textit{response pauses} (silence following a clinician's turn before the participant speaks). The same set of 7 summary statistics was calculated for each pause type present.  We utilized acoustic syllable estimation to quantify rate measures using a Praat script~\cite{deJong2009Praat} implemented in parselmouth~\cite{jadoul2018parselmouth}. Based on this count, we derived the \textit{speech rate} (syllables per second of total duration, including silence), \textit{articulation rate} (syllables per second of speaking time), and \textit{average syllable duration} (ASD).

\subsubsection{OpenSMILE eGeMAPS} Prosodic, spectral, and voice-quality features were extracted using the openSMILE eGeMAPS functional set via the openSMILE toolkit~\cite{eyben2010opensmile}. This provides a standardized set of acoustic features, including F0 (fundamental frequency) statistics, spectral balance, and MFCCs (Mel-frequency cepstral coefficients).

\subsubsection{Praat-derived prosodic and rhythmic features} Using parselmouth~\cite{jadoul2018parselmouth}, we extracted additional pitch and intensity features from the participant-only audio. We analyzed voiced frames within a physiologically plausible F0 range for adult speakers (78–350 Hz). F0 statistics were computed in Hz, log-Hz, and semitone scales. We also extracted mean intensity (dB) and RMS amplitude.

\subsubsection{pyAudioAnalysis features} Broader spectral and energy patterns were captured using pyAudioAnalysis~\cite{giannakopoulos2015pyaudioanalysis}. Mid-term (1 s) feature vectors were computed by aggregating short-term features (e.g., spectral centroid, entropy) over the entire signal. These features serve as a complementary high-dimensional representation of the acoustic texture.
\begin{table}[t] 
\centering
\caption{Feature counts for single-speaker and conversational tasks without any feature filtering.
Response pause features are not applicable to single-speaker tasks.} 
\label{tab:feature_counts}
\begin{tabular}{lrr} \hline 
\textbf{Feature group} & \textbf{Single} & \textbf{Conversation} \\\hline 
Temporal features (Utterance, Speech Rate)& 20 & 27 \\
pyAudioAnalysis & 136 & 136 \\
openSMILE eGeMAPS & 88 & 88 \\
Praat (Pitch, Intensity, Rhythm) & 11 & 11 \\
\hline
Total & 255 & 262 \\ 
\hline 
\end{tabular} 
\end{table}
\subsection{Feature filtering and exclusion criteria}
\label{filtering}

Prior to model-based feature selection, we applied a two-stage feature filtering procedure aimed at reducing features while preserving interpretability and cross-dataset consistency.

\subsubsection{Statistical filtering using the KS statistic}
As an initial univariate screening step, we applied the two-sample Kolmogorov–Smirnov (KS) test independently to each feature.
For each dataset, we computed the KS statistic ($D$), which quantifies the maximum difference between the empirical cumulative distribution functions of the cognitive impairment and control groups.
Features with a KS statistic below 0.1 were excluded as an initial filtering criterion, as higher thresholds (e.g., 0.15) resulted in excessive feature removal.

\subsubsection{Expert-guided filtering}

Following univariate statistical filtering, we applied an expert-guided exclusion step. Based on expert review from a phonetician, we excluded vowel formant--level features (e.g., \texttt{F1, F2, F3}) as they require reliable vowel- or phoneme-level alignment, while the current datasets were analyzed at the utterance-level. Furthermore, we excluded other segmental measures; while one dataset includes reading-aloud speech (i.e., Ivanova corpus),  the other corpora consist of spontaneous speech and variable word production. As such, comparisons at the segmental level (e.g., duration of voiced segments) would not be appropriate as they could be driven by the words themselves. This filtering procedure is independent of dataset-specific tuning, supporting transparent and interpretable cross-dataset comparison.

\subsection{Feature selection with Random Forest and RFECV}
After statistical and expert-guided filtering, machine learning model-based feature selection was applied using RFECV (Recursive Feature Elimination Cross Validation) with a Random Forest (RF) classifier.
RFECV iteratively removes features based on RF feature importance scores and evaluates classification performance at each elimination step using cross-validation on the training data.
The feature subset yielding the highest validation performance was retained, allowing the optimal number of features to be determined automatically.

\subsubsection{Feature configurations for comparative evaluation}

To disentangle the effects of statistical filtering, expert knowledge, and model-based feature selection, we evaluated the following five feature configurations:

\begin{itemize}
    \item \textbf{(A) All features}: No filtering; all extracted features were used.
    \item \textbf{(B) KS-only}: Features retained after univariate statistical filtering based on distributional separability.
    \item \textbf{(C) KS + Expert}: Features retained after KS-based filtering followed by expert-guided filtering.
    \item \textbf{(D) KS + RFECV}: RFECV applied on top of the KS-filtered feature set (B).
    \item \textbf{(E) KS + Expert + RFECV}: RFECV applied on top of the KS- and expert-filtered feature set (C).
\end{itemize}

\subsubsection{Evaluation protocol}

All feature configurations were evaluated using the same repeated train/test protocol.
For each dataset, we performed 20 independent runs with different random seeds, where the data were split into stratified training and test sets (80/20).
In all configurations, classification performance was evaluated on the held-out test set.

For the feature configurations (A)--(C), no model-based feature selection was applied.
In each run, a RF classifier was trained directly on the corresponding feature set and evaluated on the test split.
For the feature configurations (D) and (E), RFECV was applied exclusively within the training data of each run to determine an optimized feature subset.
A RF classifier was then trained on the selected features and evaluated on the test split.

Thus, all configurations share the same classifier, number of runs, and test-set evaluation protocol, differing only in whether RFECV is used to adaptively reduce the feature space prior to model training. We report classification performances using the area under the receiver operating characteristic curve (ROC-AUC). All filtering rules and RFECV parameters are fixed a priori and shared across datasets, enabling direct replication of the proposed pipeline.

\subsubsection{Identification of stable important features}

To identify robust acoustic biomarkers, we conducted a post-hoc stability analysis across repeated RFECV runs.
For each feature, we computed its \textit{selection frequency}, defined as the proportion of RFECV runs in which the feature was retained in the RFECV-selected feature subset that maximized cross-validated performance.
Features selected in at least 80\% of the runs (i.e., frequency $\geq 0.8$) were designated as \textit{stable important features}.
This stability-based criterion prioritizes features that are consistently informative across different data splits and reduces sensitivity to random initialization or train/test partitioning.

\section{Results and Discussion}
\subsection{Overall classification performance}
\begin{table*}[t]
\centering
\caption{Effect of feature filtering and RFECV on feature dimensionality and classification performance.
Each cell reports the number of retained features and the corresponding ROC-AUC (shown as \#Features / AUC mean (95\% confidence interval)).}
\label{tab:feature-reduction-summary}
\begin{tabular}{lccccc}
\hline
\textbf{Dataset}
& \textbf{(A) All}
& \textbf{(B) KS-only}
& \textbf{(C) KS + Expert}
& \textbf{(D) KS + RFECV}
& \textbf{(E) KS + Expert + RFECV} \\
\hline
ADReSSo
& 255 / 0.74 (0.70--0.78)
& 186 / 0.76 (0.72--0.80)
& 171 / 0.76 (0.73--0.79)
& 156 / 0.75 (0.72--0.78)
& \textbf{142 / 0.76 (0.73--0.79)} \\
Ivanova
& 255 / 0.77 (0.74--0.80)
& 199 / 0.77 (0.75--0.79)
& 178 / 0.77 (0.75--0.79)
& 121 / 0.77 (0.74--0.80)
& \textbf{114 / 0.77 (0.75--0.79)} \\
Chou
& 255 / 0.66 (0.62--0.70)
& 253 / 0.66 (0.62--0.70)
& 227 / 0.67 (0.63--0.71)
& 131 / 0.66 (0.62--0.70)
& \textbf{106 / 0.67 (0.63--0.71)} \\
PROMPT
& 262 / 0.79 (0.76--0.82)
& 251 / 0.79 (0.76--0.82)
& 227 / 0.79 (0.76--0.82)
& 156 / 0.79 (0.76--0.82)
& \textbf{101 / 0.79 (0.76--0.82)} \\
ADRC
& 262 / 0.65 (0.63--0.67)
& 202 / 0.66 (0.64--0.68)
& 180 / 0.67 (0.65--0.69)
& 102 / 0.65 (0.63--0.67)
& \textbf{\, 86 / 0.67 (0.65--0.69)} \\
\hline
\end{tabular}
\end{table*}

Table~\ref{tab:feature-reduction-summary} summarizes the joint effect of feature filtering and RFECV on both the number of retained features and classification performance across the five datasets.
Statistical filtering using the KS statistic reduced the number of retained features to varying degrees depending on dataset-specific distributional separability.
The subsequent expert-guided filtering consistently produced an additional reduction, removing features that are not appropriate at the utterance-level (e.g., formant features, segmental features).
Applying RFECV further reduced the number of features in a data-adaptive manner, yielding a smaller subset when combined with both KS filtering and expert-guided filtering.

Despite these feature reductions, classification performance measured by ROC-AUC was consistently maintained or slightly improved relative to models trained on the full feature set. In particular, the KS + Expert + RFECV configuration achieved comparable or slightly improved performance while operating on the smallest feature set. In addition, compared with KS-only filtering, the KS + Expert setting preserved performance despite additional feature reduction, indicating that expert knowledge was consistent with model-driven feature relevance. Overall, these results demonstrate that statistical filtering, expert knowledge, and model-based selection provide complementary benefits, enabling reduced, interpretable, and task-adaptive acoustic feature sets for multilingual cognitive impairment classification across languages and task modalities.

\subsection{Cross-dataset analysis of selected acoustic feature families}
\label{cross_dataset_analysis}
\begin{table*}[t]
\centering
\caption{Cross-dataset comparison of dominant acoustic features.
Top rows summarize dataset attributes, while bottom rows indicate the relative dominance
(\textit{High}, \textit{Moderate}, \textit{Low}) of each acoustic feature family within stable important features.}

\label{tab:cross_dataset_integrated}
\begin{tabular}{lccccc}
\hline
\textbf{Dataset} 
& \textbf{Ivanova} 
& \textbf{Chou} 
& \textbf{ADReSSo} 
& \textbf{ADRC} 
& \textbf{PROMPT} \\
\hline
Language & Spanish & Mandarin & English & English & Japanese \\
Task & Reading & Picture Description & Picture Description & Autobiographical Memory & Conversation \\
The number of stable important features & 53 & 23& 127& 12& 29\\
\hline

Temporal features (Duration, Speech Rate)& \textbf{High} & \textbf{High} & Moderate & \textbf{High} & Moderate \\
Pitch (F0 variability and slope)    & Moderate  & Low      & Moderate & \textbf{High} & Moderate \\
Pitch (chroma-based mean and std)  & Moderate & \textbf{High} & Moderate & Low & \textbf{High} \\
MFCC / Spectral Features (voiced; mean, std)& Moderate & Moderate & Moderate & Low & Low \\
Energy / Loudness (dynamic; mean, slope)    & Moderate & Low & Moderate & Low 
& Moderate \\
Voice quality (Jitter, Shimmer)    & Low & Low & Low & Low 
& Low \\
\hline
\end{tabular}
\end{table*}

To facilitate systematic comparison beyond dataset-specific results, we analyzed the stable important features retained by RFECV at the feature-family level across all datasets. Instead of examining individual features, this approach highlights broader acoustic categories that reveal shared and divergent patterns across languages and task modalities.
Table~\ref{tab:cross_dataset_integrated} provides a high‑level summary by listing dominant acoustic feature families for each dataset, along with their corresponding language and task (see Appendix for representative examples of stable important features for each dataset).
Dominance was determined by the proportion of stable important features belonging to each feature family,  offering a coarse-grained view of feature-family contributions.
The total number of stable important features varied substantially across datasets (e.g., larger in ADReSSo), reflecting differences in task structure, recording conditions, and feature redundancy rather than differences in how feature-families are defined.

\subsubsection{Temporal features}
Across all five datasets, temporal features were consistently selected regardless of language or task, establishing them as the most robust in our study. This cross-linguistic and cross-task consistency suggests that disruptions in speech timing constitute a core acoustic marker of cognitive decline, largely independent of linguistic structure.

Single-speaker tasks with constrained content and pacing included picture description (ADReSSo, Chou) and reading aloud (Ivanova). In these tasks, temporal characteristics were consistently captured by pause- and timing-related features, with speech rate also emerging as a stable contributor. This pattern suggests that when the utterance plan is externally anchored by a fixed stimulus or script, global articulation tempo becomes a sensitive marker of temporal disruption associated with cognitive impairment.

In contrast, speech rate features were less informative in PROMPT and ADRC, where spoken content is highly individualized. PROMPT’s naturalistic free conversations and ADRC’s autobiographical recall both require speakers to retrieve, organize, and verbalize personally variable content, introducing content‑driven variability in pacing that might obscure impairment‑related differences in speech rate. Previous research suggests that speech rate in autobiographical recall may vary with temporal distance (e.g., recent versus remote events), further increasing between-speaker variability~\cite{cohen2025pause}. As a result, temporal disruption in these settings was more robustly reflected in pause- and timing-related features, which may index diminished processing speed and working memory in individuals with cognitive impairment~\cite{hoffmann2010}.

Overall, these results show that while specific features, such as speech rate, are task-dependent, temporal features as a whole remain a universal and robust marker of CI.

\subsubsection{Pitch (F0 and chroma based) features}
In non-tonal, Indo-European languages such as English and Spanish (ADReSSo, Ivanova, ADRC), pitch dynamics derived from continuous fundamental frequency (F0) measurements were retained by RFECV as primary indicators. Although absolute F0 level measures (e.g., mean F0) were occasionally selected, the dominant F0-related features consist of variability- and dynamics-based metrics, including F0 standard deviation and slope-based features. In contrast, results using datasets of tonal or pitch-accent languages such as Mandarin Chinese (Chou) and Japanese (PROMPT), respectively, showed a clear shift towards Chroma-based features that capture relative pitch structure within an octave. 

This shift reflects the differing functional roles of pitch across phonological systems.  In non-tonal languages like English, pitch primarily serves prosodic and discourse-level functions, such as sentence-level intonation and word stress~\cite{ladefoged2015course}. Since listeners of these languages rely on pitch dynamics for prosodic information, cognitive decline may be perceived through reduced dynamic F0 control~\cite{martinez2012prosody}.

Conversely, in tonal languages like Mandarin Chinese, where lexical meaning is encoded through relative pitch categories (tones), the distribution of pitch classes (Chroma) becomes linguistically critical. Prior work has shown that Mandarin speakers tend to adopt a “holistic listening mode” that emphasizes pitch chroma proximity rather than absolute pitch height~\cite{lin2018incongruent}. This functional distinction could be one explanation as to why Chroma-based features are selected in tonal languages, as they might better capture linguistically salient pitch organization than absolute F0-based measures.

The prominence of Chroma features in Japanese (PROMPT) aligns with its characteristics as a pitch‑accent language. Although not a fully tonal language, Japanese uses word-level pitch patterns for  semantic differentiation~\cite{britannica_japanese_language}, giving lexical pitch structure greater importance than in intonation‑dominant English.  The importance placed on Chroma in the Japanese dataset could be attributable to the linguistic prominence of lexical-level pitch structure. At the same time, the comparatively weaker dominance of Chroma features in Japanese than in Mandarin can be interpreted as reflecting the more limited role of pitch in encoding lexical meaning in pitch-accent languages relative to fully tonal languages.

\subsubsection{Spectral, MFCC, energy, and voice quality features}
Unlike timing features, spectral features (such as voice quality, and MFCC-, energy--related features) exhibited strong dataset-specific behavior, limiting their cross-dataset generalizability (Table~\ref{tab:cross_dataset_integrated}). 
MFCC and spectral features were moderately important across several datasets, especially ADReSSo, but the specific retained coefficients varied widely.  Lower‑order MFCCs (1–2) offer limited interpretability, whereas higher‑order coefficients (3+) have ambiguous physical meaning, making cross‑dataset differences difficult to interpret mechanistically ~\cite{tracey2023interpretable}. These inconsistencies could reflect interactions among language, task structure, and recording conditions rather than a stable biomarker.

Energy- and loudness-related features showed strong task dependence, appearing as moderately dominant in several datasets while remaining weak in others. Given the sensitivity of intensity to recording conditions (e.g., distance from a microphone) and speaking style, their prominence may be a corpus-driven effect rather than universally diagnostic. Voice‑quality features (e.g., jitter, shimmer) showed uniformly low importance across all datasets, aligning with prior findings that they contribute little after feature selection \cite{rezaii2025voiceprints}.

Overall, MFCC, energy, and voice‑quality features function as dataset‑adaptive complements that improve within‑dataset discrimination but do not constitute robust, universal acoustic markers of cognitive decline.

\subsubsection{Task- and language-agnostic features}
We further examined which \textit{individual} acoustic features were consistently selected among the top 20 important features based on RF feature importance scores across our five datasets. Table~\ref{tab:robust_individual_features} lists the top features that appeared in three or more datasets. Consistent with our earlier findings, temporal features related to speech timing emerge as task- and language-agnostic acoustic biomarkers.

\begin{table}[t]
\centering
\caption{Most consistently selected task- and language-agnostic individual temporal features across datasets.}
\label{tab:robust_individual_features}
\footnotesize
\setlength{\tabcolsep}{4pt}
\begin{tabular}{lcc}
\hline
\textbf{Feature} & \textbf{\# Datasets} & \textbf{Family} \\
\hline
timing\_medial\_pause\_mean   & 4 & Temporal \\
pause\_ratio                 & 3 & Temporal \\
speech\_rate                 & 3 & Temporal \\
timing\_medial\_pause\_std    & 3 & Temporal \\
timing\_utterance\_median     & 3 & Temporal \\
utterance\_ratio             & 3 & Temporal \\
\hline
\end{tabular}
\end{table}

\subsection{Limitations and Future Work}
Our results reveal that temporal and language-sensitive pitch features were relatively stable across datasets, whereas spectral and energy-based features varied widely across languages and task contexts. Such variability indicates that pooling heterogeneous languages and task types can introduce competing acoustic patterns that mask disease‑related signals, helping explain the performance drops seen in prior blind pooling or transfer‑learning approaches.

However, the analysis is limited by its focus on acoustic features alone. While these capture temporal and prosodic aspects of speech, they exclude higher‑level linguistic information (semantic, syntactic, pragmatic) and non‑acoustic modalities such as visual cues, which may behave differently across languages and tasks. Incorporating lexical and semantic content represents a promising direction for future studies. Generalizability is further constrained because participants did not complete multiple task types within the same dataset, some languages were represented by only one dataset and task, and only four languages were examined. Future work should adopt multimodal frameworks that combine robust acoustic features with linguistically informed signals and draw on broader, more diverse datasets that span multiple tasks and languages to support language-agnostic cognitive impairment screening.

\section{CONCLUSIONS}
This study examined cognitive impairment-related acoustic features across structured tasks and spontaneous conversational speech in five datasets spanning four languages. Across all datasets, statistical filtering, expert-guided filtering, and model-based feature selection consistently reduced the number of features without degrading classification performance, providing stable important feature sets for feature-level analysis.
Analysis across the five datasets demonstrated that feature effectiveness is shaped by interactions between language characteristics and task.
Temporal features consistently emerged as robust markers across contexts, although their specific expression varied in a task-dependent manner.
In contrast, other acoustic features diverged more systematically: pitch-related features reflected linguistic structure of a language (tonal vs.\ non-tonal), whereas spectral-, energy-, and voice quality--related features exhibited pronounced dataset-specific behavior.
These findings caution against indiscriminate pooling of speech data across heterogeneous languages and tasks, and underscore the need for interpretable and context-aware modeling frameworks for scalable and clinically meaningful cognitive impairment screening.

\section*{Appendix}

While Section~\ref{cross_dataset_analysis} focuses on feature-family–level summaries, this appendix presents representative examples of stable important features retained by RFECV (up to ten features per dataset), facilitating interpretability and cross-dataset comparison. The full list is available at \url{https://doi.org/10.17605/OSF.IO/PB5JD}.

\section*{ACKNOWLEDGMENT}
This research was supported by JST ASPIRE (Adopting Sustainable Partnerships for Innovative Research Ecosystem: Advanced International Research Promotion Program) JPMJAP2326~\cite{aspire}, 
AMED (Japan Agency for Medical Research and Development) under Grant Number JP18he1102004, NIH grant RF1AG091509 (J.M.J. Vonk), and in part by NIH grant P30AG010129, which supported the collection of the speech sample data. This work was also supported by CITRIS and the Banatao Institute at the University of California and a UC Davis Academic Federation Grant.

\end{document}